\documentclass[runningheads]{llncs}
\usepackage[T1]{fontenc}
\usepackage{graphicx,verbatim}
\usepackage{amsfonts}
\usepackage{amsmath}
\usepackage{multirow}
\usepackage{booktabs}
\usepackage{array}
\usepackage[table]{xcolor}
\usepackage{makecell}
\usepackage{pifont}
\usepackage{marvosym}

\begin{document}
\title{Low-Rank Velocity Fields as a Structural Prior for Unsupervised 4D Medical Image Interpolation}
\titlerunning{Low-Rank Velocity Priors for 4D Medical Interpolation}
%

\author{Haojin Li\inst{1} \and
Hengzhuo Wang\inst{2,3} \and
Chang Liu\inst{2} \and
Zhiheng Ma\inst{4} \and
Heng Li\inst{2,5}\textsuperscript{\Letter} \and
Jiang Liu\inst{1,5}\textsuperscript{\Letter}}
\authorrunning{Haojin Li, Hengzhuo Wang, Chang Liu et al.}
\institute{Department of Computer Science and Engineering, Southern University of Science and Technology, Shenzhen, China\and
Faculty of Biomedical Engineering, Shenzhen University of Advanced Technology, Shenzhen, China\and
School of Biomedical Engineering, Shenzhen University Medical School, Shenzhen University, Shenzhen, China\and
Faculty of Computility Microelectronics, Shenzhen University of Advanced Technology, Shenzhen, China\and
Research Institute of Trustworthy Autonomous Systems, Southern University of Science and Technology, Shenzhen, China\\
\email{\textsuperscript{\Letter} Corresponding authors: liheng@suat-sz.edu.cn, liuj@sustech.edu.cn}
}

  
\maketitle              
\begin{abstract}
Endpoint-only unsupervised 4D medical image interpolation synthesizes intermediate volumes from sparsely sampled sequences with only the start and end volumes available for training; however, this weakly constrained setting often yields intermediates with unstable boundaries and non-physiological motion, limiting interpretability and downstream analysis. We propose low-rank velocity fields as a structural prior, constraining motion to a structured Tucker low-rank velocity field space that decomposes motion into globally shared spatial bases and a compact sample-specific core, thereby encouraging spatially correlated, anatomy-consistent deformation while suppressing voxel-wise high-frequency artifacts. To capture global coordination and local non-rigid details, we model motion in a coarse-to-fine multi-scale scheme and compose scale-wise deformations at inference to synthesize volumes at arbitrary times. We further provide a theoretical analysis showing that, under Tucker parameterization, low-rank parameters control the smoothness energy of the velocity field, explaining why low-rank modeling promotes smoother motion. Experiments on ACDC and 4D-Lung demonstrate state-of-the-art performance, remaining competitive with methods trained with intermediate-frame supervision, and producing intermediates with improved structural coherence and more stable anatomical contours.


\end{abstract}

\section{Introduction}

4D medical imaging captures anatomical dynamics and supports functional assessment, therapy monitoring, and longitudinal disease tracking~\cite{yan2025online}. However, practical acquisition constraints often lead to sparse and irregular temporal sampling, motivating 4D interpolation to synthesize intermediate volumes and increase temporal resolution for downstream analysis such as segmentation~\cite{perrin2025super} and reconstruction~\cite{li2025unsupervised}. Here, we study endpoint-only unsupervised 4D medical interpolation, where training provides only two endpoint volumes.

Prior volumetric interpolation methods are broadly motion-driven, warping endpoints via estimated deformations (or flows)~\cite{guo2020spatiotemporal,wang2025canfields,kim2024data}, or synthesis-driven, directly generating intermediate volumes~\cite{you2025fb,zhang2025temporal}, and are sometimes combined~\cite{kim2022diffusion,chen2024ultrasound}. For medical imaging, plausible intermediates must preserve anatomical continuity and topology while evolving smoothly~\cite{zhi2023coarse}. Under endpoint-only supervision, fully voxel-wise, high-dimensional motion parameterizations can overfit locally inconsistent high-frequency motion, leading to boundary instability, local misalignment, and non-physiological warping that degrades interpretability and downstream usability.

Anatomical motion is strongly spatially correlated and partially compressible: coordinated global patterns dominate, while local non-rigid details remain residual variation~\cite{pramanik2020deep,huttinga2021real}. This motivates an anatomy-aligned, compact motion representation rather than increased voxel-level freedom; thus we adopt structured low-rank motion modeling to suppress spurious high-frequency motion and improve anatomical consistency and temporal coherence of synthesized intermediates.

Two challenges arise when introducing structured low-rank motion into endpoint-only unsupervised 4D interpolation: (1) how to impose a compact, low-dimensional motion parameterization that captures shared spatial coordination yet adapts to each sample under weak supervision; (2) how to model motion across scales, as coarse global trends and fine non-rigid details call for a multi-scale representation and refinement~\cite{tian2022multi}.

To tackle these challenges, we propose a framework for endpoint-only unsupervised 4D medical image interpolation that introduces low-rank velocity fields as a structural prior. The framework separates shared coordinated motion from sample-specific variation and organizes motion modeling in a multi-scale hierarchy for coarse-to-fine refinement, thereby guiding endpoint-only learning toward anatomically plausible dynamics. Our contributions are summarized as follows:
\begin{itemize}
\item We present low-rank velocity fields as a structural prior for endpoint-only unsupervised 4D medical interpolation, improving anatomical consistency and temporal coherence, with analysis showing that low-rank parameterization bounds smoothness energy and suppresses high-frequency motion artifacts.
\item We introduce a Tucker-parameterized low-rank velocity representation with shared spatial bases and a compact, sample-dependent core to capture coordinated motion while retaining individual variation.
\item We propose a multi-scale low-rank motion modeling strategy that allocates capacity across resolutions for coarse-to-fine refinement from global coordination to local non-rigid deformation.
\item Experiments on ACDC and 4D-Lung show strong results, achieving the best structure-oriented scores with competitive reconstruction accuracy.
\end{itemize}

\section{Method}

\subsection{Problem Definition}

Unsupervised 4D interpolation is formulated on a spatial domain $\Omega \subset \mathbb{R}^3$, modeling an image sequence as a continuous-time function $I(t): \Omega \rightarrow \mathbb{R}$. We adopt endpoint-only training, observing only the endpoint volumes $I_0 := I(0)$ and $I_1 := I(1)$, while intermediate volumes $I_\alpha$ with $\alpha\in(0,1)$ are unavailable.

Given $(I_0,I_1)$, a model with parameters $\theta$ predicts an inter-endpoint velocity field for either ordered pair $(I_a,I_b)$ with $(a,b)\in\{(0,1),(1,0)\}$, represented on the voxel grid as a tensor $\mathcal{V}_{a\to b}$. It is converted to a deformation $\phi_{a\to b}=\mathrm{Exp}(\mathcal{V}_{a\to b})$, where $\mathrm{Exp}(\cdot)$ is the exponential map of a stationary velocity field implemented via scaling-and-squaring, and the target endpoint is reconstructed by warping $\hat I_b = I_a \circ \phi_{a\to b}$, where $\circ$ applies a deformation field to an image. Learning is driven by an endpoint reconstruction objective:
\begin{equation}
\theta^* \;=\; \arg\min_\theta \; 
\mathbb{E}_{(I_a,I_b)}
\Big[
\mathcal{L}\big(I_b,\hat I_b\big)
\;+\;
\lambda\,\mathcal{R}\big(\mathcal{V}_{a\to b}\big)
\Big],
\end{equation}
where $\mathcal{L}(\cdot,\cdot)$ is an endpoint similarity measure (e.g., normalized cross-correlation; NCC), and $\mathcal{R}(\cdot)$ regularizes the predicted velocity to promote plausible motion.

\begin{figure}[tp]
\includegraphics[width=\textwidth]{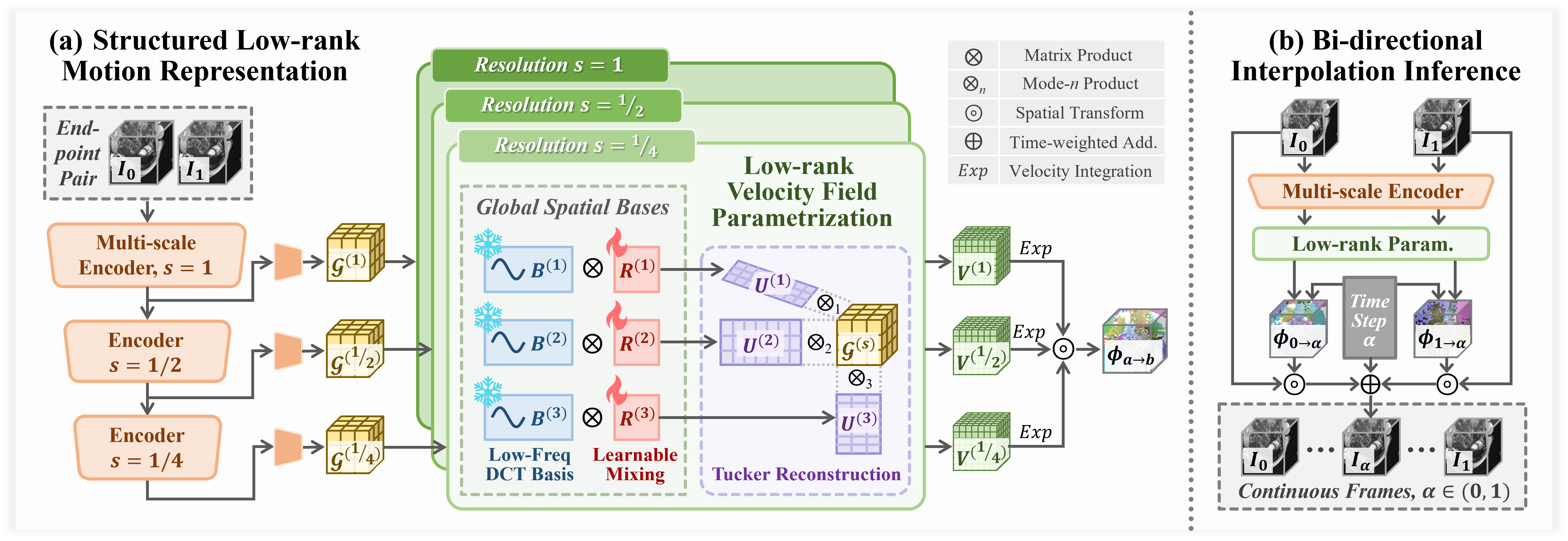}
\caption{Overview of the proposed framework. (a) Multi-scale endpoint features are projected into a structured low-rank velocity representation across resolutions via Tucker decomposition, aligning motion modeling with the coordinated and compact nature of anatomical dynamics. (b) Inference performs bi-directional continuous-time interpolation to produce structurally consistent intermediate volumes.}
\label{fig:framework}
\end{figure}

\subsection{Structured Low-rank Motion Representation}

\noindent{\textbf{Tucker Parameterization of Velocity Fields}:} 
To model high-dimensional non-rigid motion stably under endpoint-only training (Fig.~\ref{fig:framework}(a)), we constrain the predicted velocity to a structured low-rank subspace instead of a fully voxel-wise free form. A 3D velocity field on $\Omega \subset \mathbb{R}^3$, denoted $v(x)=(v_x(x),v_y(x),v_z(x))$ for $x\in\Omega$, is represented on the voxel grid as a fourth-order tensor $\mathcal{V}\in\mathbb{R}^{H\times W\times D\times 3}$. In what follows, $\mathcal{V}$ refers to this voxel-grid tensor form when defining the low-rank parameterization and regularization.

We decompose $\mathcal{V}$ into globally shared spatial bases and a compact core of combination coefficients~\cite{kolda2009tensor,de2000multilinear}:
\begin{equation}
\mathcal{V}
\;=\;
\mathcal{G}
\times_1 U^{(1)}
\times_2 U^{(2)}
\times_3 U^{(3)},
\end{equation}
where $\mathcal{G}\in\mathbb{R}^{r\times r\times r\times 3}$ is the low-dimensional core tensor, $U^{(1)}\in\mathbb{R}^{H\times r}$, $U^{(2)}\in\mathbb{R}^{W\times r}$, and $U^{(3)}\in\mathbb{R}^{D\times r}$ are spatial bases along the three axes, and $\times_n$ denotes the mode-$n$ tensor--matrix product. The bases capture domain-consistent, globally shared deformation structure, while $\mathcal{G}$ provides sample-dependent coefficients; thus, spatial correlations are absorbed by the shared bases and instance variation is governed by the compact core.

To stably construct the low-rank subspace, each spatial basis uses a fixed analytic DCT basis with learnable mixing within the same low-frequency span:
\begin{equation}
U^{(i)} = B^{(i)} R^{(i)}_{:,1:r},
\qquad
B^{(i)}\in\mathbb{R}^{N_i\times K},\;
R^{(i)}\in\mathbb{R}^{K\times K},\;
U^{(i)}\in\mathbb{R}^{N_i\times r},
\end{equation}
where $B^{(i)}$ is a fixed truncated (low-frequency) DCT basis, $R^{(i)}$ is identity-initialized and optimized to mix DCT directions within this span, and $U^{(i)}$ is the resulting rank-$r$ spatial basis. Restricting $U^{(i)}$ to a low-frequency span serves as a band-limited structural prior that discourages voxel-wise high-frequency oscillations in the induced motion~\cite{jia2023fourier,li2025aif}. We treat $K$ as a frequency budget; if $K>N_i$, the effective span is naturally limited by $N_i$.

\noindent{\textbf{Multi-scale Motion Modeling}:}
Building on the Tucker parameterization, we introduce multi-scale modeling to capture both global and local motion by predicting the velocity field coarse-to-fine at three resolutions, $1/4$, $1/2$, and $1$. Each scale $s\in\{1/4,1/2,1\}$ outputs a scale-specific low-rank velocity tensor $\mathcal{V}^{(s)}$ with its own spatial bases and core $\mathcal{G}^{(s)}$; the Tucker rank at scale $s$ is denoted by $r^{(s)}$, so that $\mathcal{G}^{(s)}\in\mathbb{R}^{r^{(s)}\times r^{(s)}\times r^{(s)}\times 3}$. The core $\mathcal{G}^{(s)}$ is regressed from scale-specific endpoint features, and all scale-wise branches are trained jointly under a unified multi-scale objective, with scales coupled via coarse-to-fine deformation composition for progressive refinement and without explicit cross-scale feature fusion.

During training, we compute and aggregate scale-wise endpoint reconstruction and smoothness losses. Let $I^{(s)}$ denote endpoints downsampled to scale $s$. Using $\mathrm{Exp}(\cdot)$ and the warping operator defined above, we integrate the scale-wise velocity $\mathcal{V}^{(s)}_{a\to b}$ into a deformation $\phi^{(s)}_{a\to b}$ (progressively refined from coarse to fine) and warp $I^{(s)}_a$ to reconstruct $\hat I^{(s)}_b$:
\begin{equation}
\mathcal{L}_{\mathrm{total}}
=
\sum_{s} \omega^{(s)} \Big(
-\lambda_1\,\mathrm{NCC}(I^{(s)}_b,\hat I^{(s)}_b)
+
\lambda_2\,
\big\| I^{(s)}_b-\hat I^{(s)}_b \big\|_{\mathrm{C}}
+
\lambda_3\,\mathcal{R}_{H^1}(\mathcal{V}^{(s)}_{a\to b})
\Big),
\end{equation}
where $\sum_s \omega^{(s)}=1$, $\mathrm{NCC}(\cdot,\cdot)$ denotes local normalized cross-correlation, $\|\cdot\|_{\mathrm{C}}$ is the Charbonnier norm, and $\mathcal{R}_{H^1}$ penalizes velocity magnitude and spatial gradients as an additional safeguard for smooth, integrable motion. All reconstruction terms at scale $s$ are evaluated between $I^{(s)}_b$ and $\hat I^{(s)}_b$ at the same resolution.

\subsection{Inference for Continuous-Time Interpolation}

At inference (Fig.~\ref{fig:framework}(b)), we evaluate the learned multi-scale low-rank representation to synthesize intermediate volumes at arbitrary $\alpha\in(0,1)$. Given endpoints $(I_0,I_1)$, the predicted velocity tensors at three spatial scales, $\mathcal{V}^{(1/4)}, \mathcal{V}^{(1/2)}, \mathcal{V}^{(1)}$, are resampled to a common voxel grid, with magnitudes scaled according to resolution. For each direction $a\to b$, we integrate the scale-specific velocity via $\mathrm{Exp}(\cdot)$ to obtain a deformation $\phi^{(s)}_{a\to b}$, and form the endpoint deformation by coarse-to-fine composition:
\begin{equation}
\phi_{a\to b}
=
\phi^{(1)}_{a\to b}\circ \phi^{(1/2)}_{a\to b}\circ \phi^{(1/4)}_{a\to b}.
\end{equation}

Deformations are applied right-to-left across scales (coarse to fine), and each $\phi^{(s)}_{a\to b}$ is computed by integrating the scale-$s$ velocity with $\mathrm{Exp}(\cdot)$ and resampling it to the common grid before composition.

To obtain an intermediate time $\alpha$, we use both predicted directions ($0\to 1$ and $1\to 0$) and scale each stationary velocity by the corresponding travel time (proportional to $\alpha$ or $1-\alpha$) prior to exponential-map integration~\cite{vercauteren2009diffeomorphic}, yielding $\phi_{0\to\alpha}$ and $\phi_{1\to\alpha}$. The intermediate prediction is then computed by warping both endpoints and fusing them with time-based weights:
\begin{equation} \hat I_\alpha = (1-\alpha)\,\big(I_0 \circ \phi_{0\to\alpha}\big) + \alpha\,\big(I_1 \circ \phi_{1\to\alpha}\big). \end{equation}

\subsection{Theoretical Analysis of Energy Bounds}

\paragraph{\textbf{Proposition: Energy control by low-rank parameters.}}
In Tucker form, the smoothness energy of the full velocity field is upper-bounded by the core magnitude, up to a multiplicative constant determined by the smoothing operator and the spatial bases; consequently, bounding the factor norms and the core norm controls the regularity of the induced deformation~\cite{kolda2009tensor}.
Define the quadratic smoothness energy:
\begin{equation}
E(\mathcal{V}) \;:=\; \big\| L\,\mathrm{vec}(\mathcal{V}) \big\|_2^2, 
\end{equation}
where $L$ is a fixed linear smoothing/differential operator and $\mathrm{vec}(\cdot)$ stacks tensor entries into a vector.
Let $\lambda_{\max}$ be the largest eigenvalue of $L^\top L$, i.e., $\lambda_{\max}=\|L\|_2^2$.
For a Tucker-parameterized velocity tensor
$\mathcal{V}=\mathcal{G}\times_1 U^{(1)}\times_2 U^{(2)}\times_3 U^{(3)}$,
the following upper bound holds:
\begin{equation}
E(\mathcal{V})
\;\le\;
\lambda_{\max}\Bigl(\prod_{i=1}^3\|U^{(i)}\|_2\Bigr)^2\|\mathcal{G}\|_F^2 ,
\label{eq:tucker_energy_bound}
\end{equation}
where $\|\cdot\|_2$ is the matrix spectral norm and $\|\cdot\|_F$ is the Frobenius norm.

\paragraph{\textbf{Proof.}}
By the operator norm inequality,
$\|L\,\mathrm{vec}(\mathcal{V})\|_2 \le \|L\|_2\,\|\mathrm{vec}(\mathcal{V})\|_2$,
we have $E(\mathcal{V}) \le \lambda_{\max}\|\mathcal{V}\|_F^2$.
Moreover, for any mode-$n$ product, $\|\mathcal{X}\times_n U\|_F \le \|U\|_2\|\mathcal{X}\|_F$~\cite{beg2005computing}; applying this sequentially to the Tucker reconstruction yields
$\|\mathcal{V}\|_F \le (\prod_{i=1}^3\|U^{(i)}\|_2)\|\mathcal{G}\|_F$.
Combining the two bounds proves \eqref{eq:tucker_energy_bound}.

\paragraph{\textbf{Corollary (Rank scaling).}}
Under bounded factors $\|U^{(i)}\|_2\le c$ and bounded core amplitude $|\mathcal{G}_{pqr\ell}|\le \beta$ for constants $c$ and $\beta$,
the worst-case energy bound scales at most cubically with the rank: for $\mathcal{G}\in\mathbb{R}^{r\times r\times r\times 3}$, $\|\mathcal{G}\|_F^2 \le 3r^3\beta^2$, hence
\begin{equation}
E(\mathcal{V}) \;\le\; 3\,\lambda_{\max}\,c^6\,r^3\,\beta^2 . 
\label{eq:rank_bound}
\end{equation}
The same bound applies to each scale $s$ by replacing $r$ with $r^{(s)}$.

\section{Experiment}

\subsection{Experimental Settings}

\noindent \textbf{Datasets:} We evaluate on ACDC~\cite{bernard2018deep} and 4D-Lung~\cite{hugo2016data}. ACDC contains 150 cardiac MRI cases with an 80/20/50 (train/val/test) split, resized to $160\times160\times16$; end-diastolic and end-systolic phases are used as endpoints. 4D-Lung contains 500 cone-beam CT volumes with a patient-level 306/84/110 split, resized to $128\times128\times32$; 0\% and 50\% respiratory phases are used as start/end frames. All volumes are histogram-equalized and linearly normalized to $[0,1]$.

\noindent \textbf{Metrics:} We report Normalized Mutual Information (NMI), Normalized Mean Squared Error (NMSE), and Structural Similarity (SSIM) to measure statistical dependence, reconstruction error, and structural similarity of results.

\noindent \textbf{Implementation Settings:} All methods are implemented in PyTorch and trained on NVIDIA RTX A6000 GPUs under a unified protocol. We use AdamW with an initial learning rate of $2\times10^{-4}$ and a cosine annealing scheduler. Models are trained for 500/200 epochs on ACDC/4D-Lung with early stopping; batch size is 1 with gradient accumulation of 8. For our method, endpoint features are extracted at three resolutions ($1/4$, $1/2$, $1$) using lightweight scale-specific 3D convolutional encoder branches trained jointly; scale-wise motion heads refine from coarse to fine, and low-frequency bases are truncated to $K=128$. Scale weights are set to $\omega^{(1/4)}=0.2$, $\omega^{(1/2)}=0.3$, and $\omega^{(1)}=0.5$; Tucker ranks are set to $r^{(1/4)}=64$, $r^{(1/2)}=32$, and $r^{(1)}=8$. We set $\lambda_1=\lambda_2=1$ and $\lambda_3=0.05$.

\subsection{Comparison Experiment}

\begin{table}[t]
\centering
\caption{Quantitative comparison on ACDC and 4D-Lung datasets.}
\footnotesize 
\label{tab:comparison}
\begin{tabular}{lccccccccc} 
\toprule
\multirow{2}{*}{Method} & \multirow{2}{*}{Year} && \multicolumn{3}{c}{ACDC} & & \multicolumn{3}{c}{4D-Lung} \\ 
\cmidrule(r){4-6} \cmidrule(l){7-10}
& && NMI$_{\times 10^{-2}}^{\uparrow}$ & NMSE$_{\times 10^{-3}}^{\downarrow}$ & SSIM$_{\times 10^{-2}}^{\uparrow}$ 
& & NMI$_{\times 10^{-2}}^{\uparrow}$ & NMSE$_{\times 10^{-3}}^{\downarrow}$ & SSIM$_{\times 10^{-2}}^{\uparrow}$ \\ 
\midrule
VM & 2019 && 53.54$_{\pm 0.37}$ & 19.18$_{\pm 0.34}$ & 94.68$_{\pm 0.10}$ 
& & 43.98$_{\pm 0.36}$ & 39.02$_{\pm 0.60}$ & 86.06$_{\pm 0.29}$ \\
TM & 2022 && 64.63$_{\pm 0.67}$ & 14.43$_{\pm 0.62}$ & 96.64$_{\pm 0.12}$ 
& & 48.95$_{\pm 0.10}$ & 34.96$_{\pm 0.40}$ & 84.78$_{\pm 0.12}$ \\
SVIN & 2020 && 60.18$_{\pm 0.33}$ & 9.85$_{\pm 0.89}$ & 97.35$_{\pm 0.15}$ 
& & 43.17$_{\pm 0.39}$ & 30.28$_{\pm 2.03}$ & 85.44$_{\pm 0.29}$ \\
UVINet & 2024 && 72.25$_{\pm 0.17}$ & 4.81$_{\pm 0.12}$ & 98.62$_{\pm 0.03}$ 
& & 55.16$_{\pm 0.84}$ & 20.93$_{\pm 1.27}$ & 88.75$_{\pm 0.67}$ \\
\rowcolor{gray!15}DDM & 2022 && 64.14$_{\pm 1.35}$ & 10.36$_{\pm 0.30}$ & 97.24$_{\pm 0.87}$ 
& & 42.54$_{\pm 0.89}$ & 33.19$_{\pm 0.73}$ & 85.23$_{\pm 0.27}$ \\
\rowcolor{gray!15}LDDM & 2024 && 72.46$_{\pm 0.30}$ & 4.88$_{\pm 0.04}$ & 98.56$_{\pm 0.01}$ 
& & 56.58$_{\pm 8.08}$ & 28.62$_{\pm 0.80}$ & 83.50$_{\pm 6.09}$ \\
\rowcolor{gray!15}FB-Diff & 2025 && 61.33$_{\pm 0.21}$ & 8.64$_{\pm 0.21}$ & 97.29$_{\pm 0.09}$ 
& & 49.28$_{\pm 1.34}$ & 27.39$_{\pm 6.26}$ & 80.91$_{\pm 3.71}$ \\
\rowcolor{gray!15}TMSDF & 2025 && 66.40$_{\pm 0.83}$ & \textbf{4.72}$_{\pm 0.23}$ & 98.56$_{\pm 0.07}$ 
& & 48.36$_{\pm 0.11}$ & \textbf{16.85}$_{\pm 0.09}$ & 90.26$_{\pm 1.09}$ \\
\midrule
\textbf{Ours} & - && \textbf{72.61}$_{\pm 0.13}$ & 4.74$_{\pm 0.05}$ & \textbf{98.62}$_{\pm 0.02}$ 
& & \textbf{62.34}$_{\pm 0.20}$ & 17.00$_{\pm 0.43}$ & \textbf{91.29}$_{\pm 0.13}$ \\ 
\bottomrule
\end{tabular}
\footnotesize \colorbox{gray!15}{\strut\,Gray background\,}: methods trained with intermediate-frame supervision.
\end{table}

We compare against classical unsupervised deformation models (Voxelmorph (VM)~\cite{balakrishnan2019voxelmorph}, Transmorph (TM)~\cite{chen2022transmorph}), recent state-of-the-art unsupervised interpolation networks (SVIN~\cite{guo2020spatiotemporal}, UVINet~\cite{kim2024data}), and representative methods trained with intermediate-frame supervision (DDM~\cite{kim2022diffusion}, LDDM~\cite{chen2024ultrasound}, FB-Diff~\cite{you2025fb},
TMSDF~\cite{zhang2025temporal}).

The quantitative results in Tab.~\ref{tab:comparison} reveal distinct performance patterns across the evaluated methods. While recent unsupervised interpolation models perform strongly on ACDC, they exhibit a notable gap on 4D-Lung due to the challenge of preserving globally coordinated respiratory motion. Conversely, methods utilizing intermediate-frame supervision often reduce reconstruction error, yet their NMI/SSIM scores do not consistently improve, confirming that appearance gains do not necessarily translate into stronger anatomical correspondence. Our approach achieves leading NMI/SSIM on both datasets, particularly on 4D-Lung, while maintaining competitive reconstruction error. Despite relying only on endpoint supervision, it reaches supervised-level overall quality and yields more anatomically consistent, temporally coherent intermediates.

\begin{figure}[tp]
\includegraphics[width=1\textwidth]{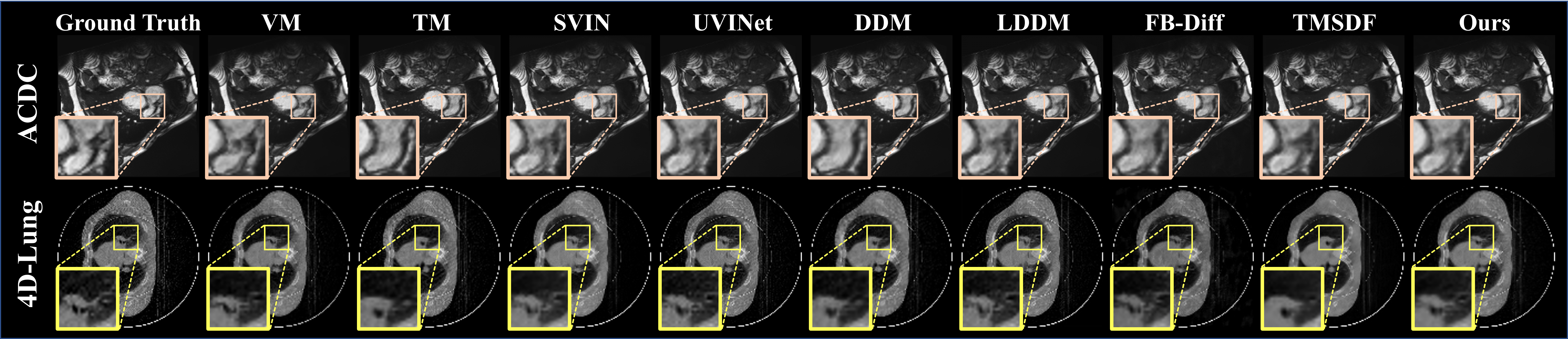}
\caption{Qualitative comparison on ACDC and 4D-Lung datasets.
}
\label{fig:comparison}
\end{figure}

Fig.~\ref{fig:comparison} shows qualitative comparisons. On ACDC, our method produces more regular contours and stable anatomy, especially in complex regions; on 4D-Lung, it better preserves coherent global organization. Baselines often exhibit local inconsistencies, while diffusion-style methods may improve appearance but compromise structural stability, yielding more anatomically consistent and visually stable intermediates.

\subsection{Ablation Study}

\begin{figure}[!t]
\includegraphics[width=1\textwidth]{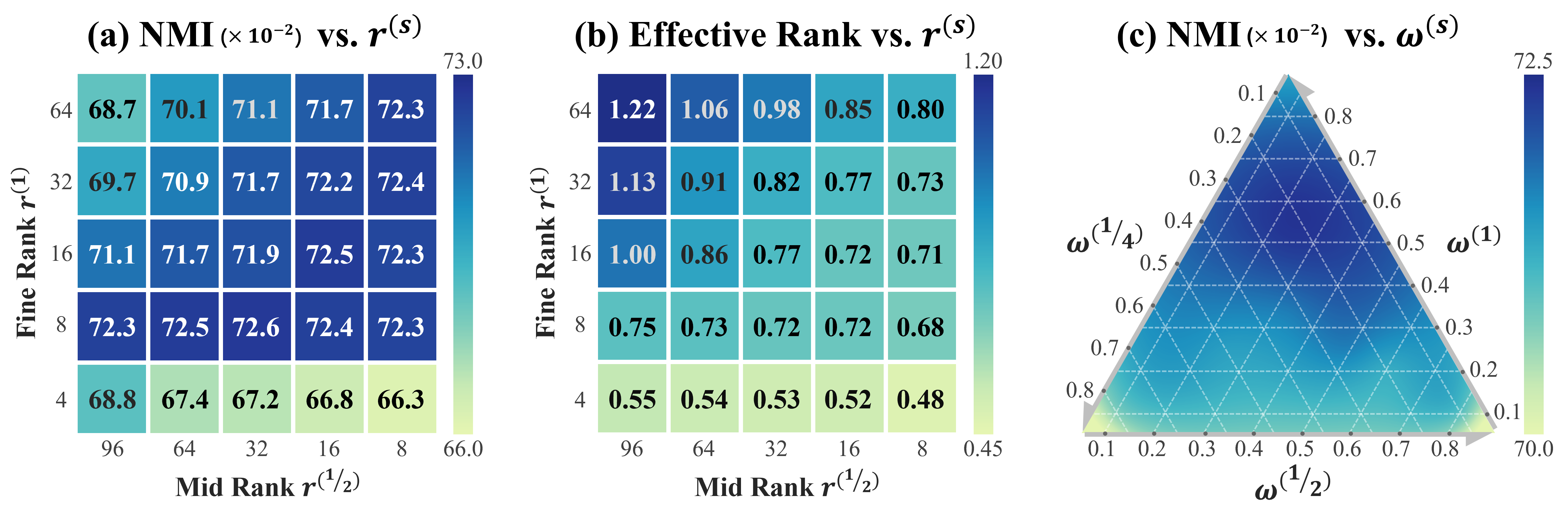}
\caption{Ablation on ACDC. (a) NMI under different Tucker rank settings. (b) Velocity field Effective Rank~\cite{park2024self} (log$_{10}$-scaled). (c) NMI vs.\ scale weights (ternary diagram).}
\label{fig:ablation}
\end{figure}

\noindent{\textbf{Effect of Multi-scale Rank}:}
As shown in Fig.~\ref{fig:ablation}(a), we fix the coarse Tucker rank (given its low sensitivity) and vary the mid/fine ranks, reporting NMI. Increasing Tucker rank improves interpolation quality in low-capacity regimes, while the gain saturates once the fine-scale rank becomes large. Fig.~\ref{fig:ablation}(b) reports the Effective Rank of the resulting velocity fields (computed following~\cite{park2024self}) as a proxy for motion complexity: lower Effective Rank corresponds to simpler, more compressible motion, and it increases monotonically with Tucker rank in our setting. Performance is best at low-to-moderate Effective Ranks, whereas overly small ranks underfit and degrade accuracy. Overall, Tucker rank offers a direct knob to tune motion complexity via the induced Effective Rank, and both under-constraint and excessive capacity can lead to suboptimal results.

\noindent{\textbf{Effect of Scale Weights}:}
We ablate the scale-wise loss weights in Fig.~\ref{fig:ablation}(c) by enforcing $\sum_s \omega^{(s)}=1$ and visualizing NMI on a ternary plot. Overall, allocating more weight to the fine scale tends to improve performance, but overly extreme fine-dominant settings degrade NMI; the best region favors a fine-leaning yet balanced combination that retains non-negligible mid/coarse contributions. This indicates that fine-scale supervision is most effective for refinement, while mid/coarse guidance helps preserve global coherence and prevent locally driven inconsistencies, yielding the most reliable training signal across scales.

\begin{table}[!t]
\centering
\caption{Loss term ablation on ACDC.}
\label{tab:ablation}
\begin{tabular}{%
>{\centering\arraybackslash}m{1.1cm}
>{\centering\arraybackslash}m{1.1cm}
>{\centering\arraybackslash}m{1.1cm}
|
>{\centering\arraybackslash}m{1.8cm}
>{\centering\arraybackslash}m{1.8cm}
>{\centering\arraybackslash}m{1.8cm}
}
\toprule
NCC & Charb & Reg & NMI$_{\times 10^{-2}}^{\uparrow}$ & NMSE$_{\times 10^{-3}}^{\downarrow}$ & SSIM$_{\times 10^{-2}}^{\uparrow}$ \\
\midrule
$\checkmark$ & & $\checkmark$ & 72.22 & \textbf{4.69} & \textbf{98.63} \\
& $\checkmark$ & $\checkmark$ & 72.35 & 4.81 & 98.59 \\
$\checkmark$ & $\checkmark$ & & 71.45 & 5.18 & 98.49 \\
\midrule
$\checkmark$ & $\checkmark$ & $\checkmark$ & \textbf{72.61} & 4.74 & 98.62 \\
\bottomrule
\end{tabular}
\end{table}

\noindent{\textbf{Loss-Term Ablation}: }
Tab.~\ref{tab:ablation} ablates the loss by enabling/disabling NCC, the Charbonnier term, and the regularizer. Dropping Charbonnier can slightly improve NMSE/SSIM but may destabilize optimization, suggesting it complements NCC by tolerating local intensity deviations under unsupervised reconstruction. Regularization is also beneficial: without it, the model tends to produce larger and noisier motions, whereas enabling it stabilizes the scale of core tensors and curbs excessive growth of low-rank parameters, yielding smoother fields and more robust training.

\section{Conclusion}

We proposed a structured low-rank framework for unsupervised 4D medical image interpolation. By parameterizing velocity fields with a Tucker low-rank form and modeling motion in a coarse-to-fine multi-scale manner, the method favors spatially correlated, anatomy-consistent deformation under endpoint-only training. Our analysis links the smoothness energy of the velocity field to low-rank parameters, offering theoretical support for the structural prior. Experiments on ACDC and 4D-Lung show best performance on structure-oriented metrics with competitive interpolation error.

\bibliographystyle{splncs04}
\bibliography{refer}

\end{document}